\documentclass[lettersize,journal]{IEEEtran}
\usepackage{amsmath,amsfonts}
\usepackage{array}
\usepackage{textcomp}
\usepackage{stfloats}
\usepackage{url}
\usepackage[colorlinks,linkcolor=red]{hyperref}
\usepackage{verbatim}
\usepackage{graphicx}
\usepackage{booktabs}
\usepackage{cite}
\usepackage{multicol}
\usepackage{algorithm}
\usepackage{lipsum} 
\usepackage{caption}
\usepackage{longtable}
\usepackage{hyperref}
\usepackage{xcolor}
\usepackage{color}
\usepackage{bbm, dsfont}
\usepackage{multirow}
\usepackage{amsfonts}
\usepackage[noend]{algpseudocode}
\usepackage{tikz,xcolor}

\usepackage[switch]{lineno}

\definecolor{lime}{HTML}{A6CE39}
\DeclareRobustCommand{\orcidicon}{
\begin{tikzpicture}
\draw[lime, fill=lime] (0,0)
circle[radius=0.16]
node[white]{{\fontfamily{qag}\selectfont \tiny \.{I}D}};
\end{tikzpicture}
\hspace{-2mm}
}
\foreach \x in {A, ..., Z}{%
\expandafter\xdef\csname orcid\x\endcsname{\noexpand\href{https://orcid.org/\csname orcidauthor\x\endcsname}{\noexpand\orcidicon}}
}

\begin{document}
\title{SIAVC: Semi-Supervised Framework for Industrial Accident Video Classification}

\author{ 
        Zuoyong Li\hspace{-1.5mm}\orcidA{},~\IEEEmembership{Member, IEEE,}
        Qinghua Lin\hspace{-1.5mm}\orcidB{},~\IEEEmembership{Student Member, IEEE,}
        Haoyi Fan\hspace{-1.5mm}\orcidC{},~\IEEEmembership{Member, IEEE,}
        Tiesong Zhao\hspace{-1.5mm}\orcidD{},~\IEEEmembership{Senior Member, IEEE }
        \hspace{-1.6mm}and David Zhang\hspace{-1.5mm}\orcidE{},~\IEEEmembership{Life Fellow, IEEE}

\thanks{This work is partially supported by the National Key Research and Development Program of China (Grant No. 2022YFC3302200), National Natural Science Foundation of China (61972187, 62276146), and Key Project of Colleges and Universities of Henan Province (23A52002). (Corresponding author: Haoyi Fan.)
 }
\thanks{Zuoyong Li is with the Fujian Provincial Key Laboratory of Information Processing and Intelligent Control, College of Computer and Control Engineering, Minjiang University, 350121, Fuzhou, China (email: fzulzytdq@126.com)}
\thanks{Qinghua Lin is with the School of Computer Science and Mathematics, Fujian University of
Technology, 350118, Fuzhou, China (email: Akametris@163.com)}
\thanks{Haoyi Fan is with the School of Computer and Artificial Intelligence, Zhengzhou University, 45000, Zhengzhou, China (email: fanhaoyi@zzu.edu.cn)}
\thanks{Tiesong Zhao is  with the Fujian Key Lab for Intelligent Processing and Wireless Transmission of Media Information, Fuzhou University, 350108, Fuzhou, China.  (email: t.zhao@fzu.edu.cn)}
\thanks{David Zhang is with the School of Data Science, The Chinese University of Hong Kong (Shenzhen), 518172, Shenzhen, China (e-mail:davidzhang @cuhk.edu.cn).}
}

\markboth{Journal of \LaTeX\ Class Files,~Vol.~14, No.~8, August~2021}%
{Shell \MakeLowercase{\textit{et al.}}: A Sample Article Using IEEEtran.cls for IEEE Journals}


\maketitle

\begin{abstract} 
Semi-supervised learning suffers from the imbalance of labeled and unlabeled training data in the video surveillance scenario. In this paper, we propose a new semi-supervised learning method called SIAVC for industrial accident video classification. Specifically, we design a video augmentation module called the Super Augmentation Block (SAB). SAB adds Gaussian noise and randomly masks video frames according to historical loss on the unlabeled data for model optimization.
Then, we propose a Video Cross-set Augmentation Module (VCAM) to generate diverse pseudo-label samples from the high-confidence unlabeled samples, which alleviates the mismatch of sampling experience and provides high-quality training data. 
Additionally, we construct a new industrial accident surveillance video dataset with frame-level annotation, namely ECA9, to evaluate our proposed method. 
Compared with the state-of-the-art semi-supervised learning based methods, SIAVC demonstrates outstanding video classification performance, achieving 88.76\% and 89.13\% accuracy on ECA9 and Fire Detection datasets, respectively. 
The source code and the constructed dataset ECA9 will be released in \url{https://github.com/AlchemyEmperor/SIAVC}.

\end{abstract}

\begin{IEEEkeywords}
Video classification, consistency regularization, distribution alignment, deep learning.
\end{IEEEkeywords}

\section{Introduction}
\IEEEPARstart{F}{ire} is one of the most destructive accidents, causing countless casualties and property losses \cite{FP15}. With the widespread adoption of video surveillance, researchers are dedicated to monitoring fire accidents, especially in industrial and warehouse facilities, yielding encouraging results. However, more accidents in production environments need attention, such as leaks, vehicle collisions, and blocked fire exits, potentially adversely affecting industrial productivity and personnel and property safety.

\begin{figure}[!t]
  \centering
  \includegraphics[width=0.5\textwidth]{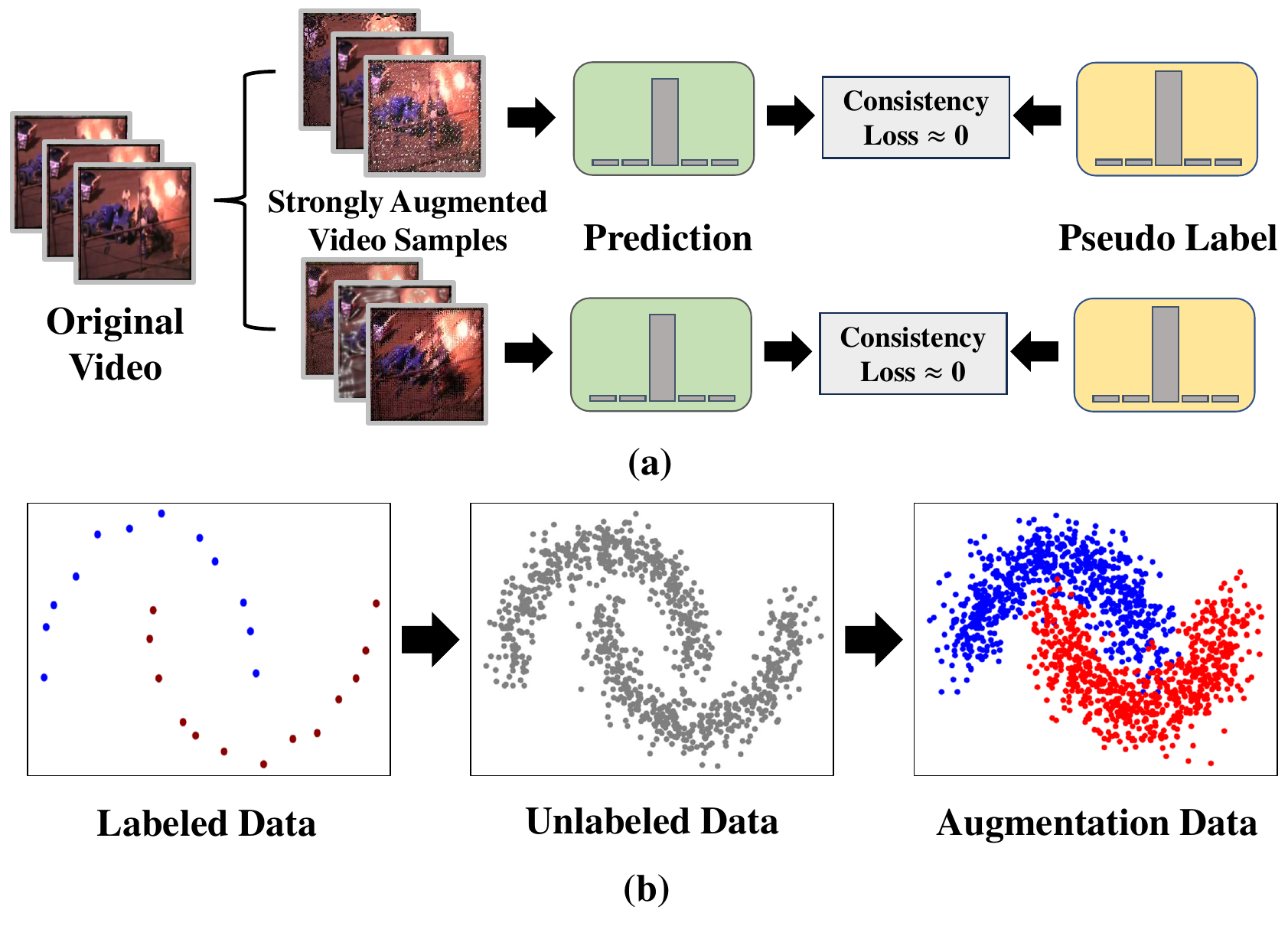}
  \caption{Motivation of the proposed method. (a) The model's high confidence with strongly augmented samples results in a consistency loss close to zero. (b) Interpolation between labeled and unlabeled data to generate pseudo-label data.}
  \label{Intro}
\end{figure}

The computer vision-based video accident detection method has received widespread attention from researchers due to its comprehensive coverage, low deployment cost, and quick response time. However, the restricted labeled data limits the vision-based model's performance. Moreover, constructing a high-quality video dataset with manual labels for video accidents remains costly. 
Nevertheless, researchers have embarked on various promising approaches, such as weak, semi, and unsupervised methods, to reduce reliance on high-quality annotated data. Semi-supervised learning aims to utilize a portion of labeled data and a large amount of unlabeled data to achieve a model's performance on par with even surpassing supervised learning.

Semi-supervised learning approaches for classification include consistency regularization and adversarial distribution alignment. The former adds varying degrees of perturbation to unlabeled data, achieved through strong and weak augmentations of the same sample. It encourages the model to produce the same output distribution and learn robust feature representations from unlabeled data \cite{SKF20}. The latter employs data augmentation strategies to obtain diverse pseudo-label samples and reduce the mismatch in sampling empirical distributions \cite{WFR22}. The study \cite{GGE23} introduce the concept of ``naive samples," indicating that in consistency regularization, a portion of strongly augmented samples can still be learned well by the model, even with significant perturbations. The ``naive samples" results in the model not benefiting from these strongly augmented samples. Additionally, the number of labeled samples heavily influences adversarial distribution alignment. Having too few labeled samples makes the generated augmented sample patterns monotonous and even hinders the model's optimization. The motivations of the proposed method, as shown in Fig. \ref{Intro}, and the limitations of previous work can be summarized as follows: (1) Model optimization has limited benefits from strongly augmented unlabeled samples.  (2) The pseudo-label samples generated through augmentation are monotonous when labeled data is scarce.

To address these issues, we propose a \underline{\textbf{S}}emi-supervised framework for \underline{\textbf{I}}ndustrial \underline{\textbf{A}}ccident \underline{\textbf{V}}ideo \underline{\textbf{C}}lassification called SIAVC. To ensure the strongly augmented unlabeled samples continue promoting model optimization, we propose SAB for further augmentation of samples. Inspired by \cite{GGE23}, for each strongly augmented sample of unlabeled data, we retain its historical consistency loss and use OTSU \cite{ONA79} to obtain a loss threshold. Then, we add random masking and Gaussian noise to each frame when the current strong augmented sample's loss is lower than the threshold.
Furthermore, we design VCAM to generate diverse pseudo-label samples. VCAM incorporates unlabeled samples with high confidence into the augmentation queue. After random shuffling, labeled and unlabeled data generate pseudo-label data through one-to-one interpolation, expand training data, and alleviate the issue of sampling experience mismatch. 
Additionally, we construct a dataset for the classification of nine types of accidents in hub-level express scenarios, including conveyor belt dropping, conveyor belt malfunction, damaged parcels, stacking explosion, blocked fire exits, leaks, fires, transport vehicle collisions, and improper placement.
The provided dataset includes video-level annotations for hub-level express scenario accident videos, with frame-level anomaly annotations provided. Through the proposed semi-supervised learning framework, SIAVC exhibits outstanding video classification performance compared to several state-of-the-art methods on publicly available fire accident datasets and the newly constructed dataset.

Our contributions are summarized as follows:

\begin{itemize}
\item 
We propose a Video Cross-set Augmentation Module (VCAM) by incorporating high-confidence unlabeled samples into the augmentation queue and generating various pseudo-label samples through interpolation, to mitigate the sampling experience mismatch while expanding the training data.
\end{itemize}

\begin{itemize}
\item 
We propose a Super Augmentation Block (SAB) that adds random mask and Gaussian noise to frames based on historical losses to re-augment high-confidence samples. SAB allows for the re-utilization of these strongly augmented samples for better consistency regularization.
\end{itemize}

\begin{itemize}
\item 
We propose a multi-class accident video dataset called Express Center Accidents 9 (ECA9). ECA9 comprises nine typical accidents in hub-level express processing centers, and provides video-level labels and frame-level anomaly labels. To the best of our knowledge, ECA9 is the first surveillance video dataset for industrial accident scenes.
\end{itemize}

\begin{figure*}[!t]
  \centering
  \includegraphics[width=0.95\textwidth]{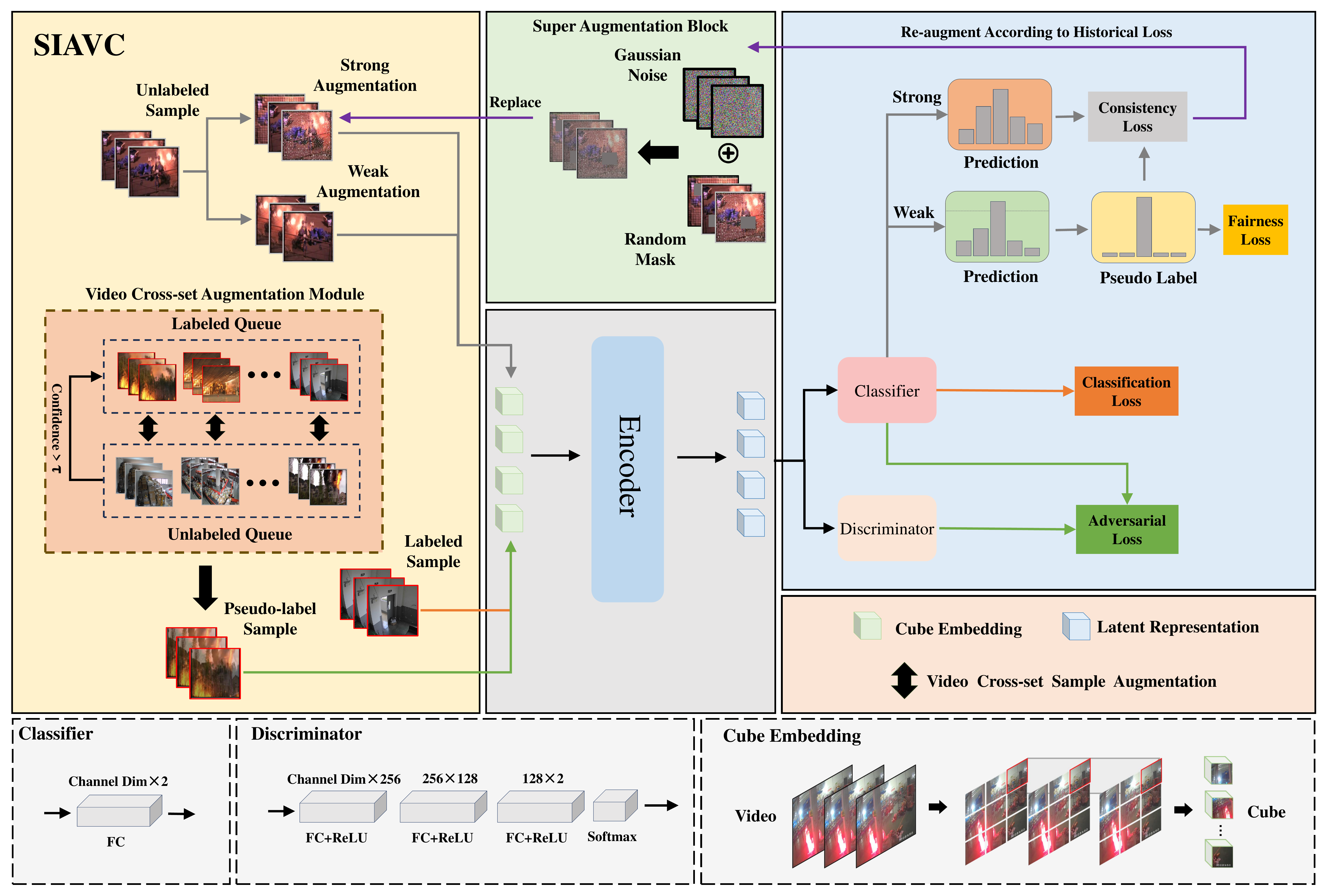}
  \caption{Overview of SIAVC. We first augment unlabeled samples to obtain their weakly augmented and strongly augmented counterparts. Then, we use VCAM to interpolate between unlabeled and labeled samples that generate pseudo-label samples. After cube embedding and sending to the encoder, the classifier and discriminator outputs predictions of these samples. Next, we compute consistency and fairness loss for predictions on strongly and weakly augmented samples. We update classification loss for labeled samples and adversarial loss for pseudo-label samples. Besides, we use SAB to re-augment these strongly augmented samples according to historical loss and consistency loss for the next iteration.}
  \label{OverView}
\end{figure*}

\section{Related Work}
\subsection{Video Classification}
Early video recognition heavily relied on manually designed features. With the advancement of deep learning, video recognition tasks based on 3D convolutional neural networks and Transformers have seen significant development. For instance, \textit{Tran et al.} \cite{TDV19} designs a 3D channel-separated convolution, which provides a form of regularization by separating channel interactions and spatiotemporal interactions, thereby increasing classification accuracy while reducing computational costs. \textit{Chadha et al.} \cite{CAV19} design a two-stream convolutional neural network that extracts information from compressed bitstreams to achieve video classification. OTAM \cite{CKF20} effectively leverages temporal information in video data through ordered temporal alignment, improving few-shot video classification. VideoSSL \cite{JLV21} simultaneously trains 2D and 3D convolutional neural networks, using confidence from unlabeled data and normalized probabilities predicted by image classifier as adjustment signals to capture information about objects of interest in the video. \textit{Li et al.} \cite{LYM22} achieves a unified architecture for image and video classification, as well as object detection, by combining decomposed relative position embeddings and residual pooling connections. \textit{Wu et al.} \cite{WWR23} focus on knowledge transfer in video classification tasks, achieving effective knowledge transfer by replacing the classifier with different knowledge from pre-trained models.

In recent years, deep learning-based video classification tasks have shown promising prospects, but these efforts still rely heavily on a substantial amount of labeled data to achieve accurate video classification. However, annotating video data is a time-consuming and labor-intensive process, making it often challenging to obtain a large volume of labeled data. Semi-supervised learning aims to harness a portion of labeled data along with a significant amount of unlabeled data to achieve model performance comparable to supervised learning.

\subsection{Semi-supervised Classification}
Recently, researchers have shown interest in semi-supervised learning because of its limited reliance on labeled data. Among the various approaches explored, pseudo-label has emerged as a mature technique in semi-supervised classification. The fundamental concept of pseudo-label is to train an initial model on labeled samples to make predictions for unlabeled samples, then retrain the model using the predicted results as labels for the unlabeled samples. Several variants of pseudo-label have been proposed to enhance its effectiveness. Pseudo-Label \cite{LDH13} utilizes pseudo-labels generated from unlabeled data during model training. UPS \cite{RMN21} tackles the problem of inaccurate high-confidence predictions by minimizing noise during training with an uncertainty-aware pseudo-label selection method, improving the precision of pseudo-labels. MixMatch \cite{BD19} combines labeled and unlabeled data as augmented data, assigning low-entropy labels to the unlabeled augmented data. FixMatch \cite{SKF20} generates high-confidence pseudo-labels for weakly augmented data and trains the model on their strongly augmented versions using these pseudo-labels. FlexMatch \cite{ZB21} introduces Curriculum Pseudo Labeling, which adjusts the thresholds of different classes to incorporate informative unlabeled data and their pseudo-labels flexibly. 
FreeMatch \cite{WY22} adapts the confidence threshold of pseudo-labels according to the learning state by introducing an adaptive class fair regularization penalty. SoftMatch \cite{CH23} addresses the quantity-quality trade-off issue in thresholded pseudo-labeling methods by weighting confidence samples during training to maintain high-quality pseudo-labels. SimMatch \cite{ZM22} applies consistency regularization at both semantic and instance levels, promoting consistent class predictions among different augmented views of the same instance and establishing similar relationships with other instances. SelfMatch \cite{KB21} combines contrastive self-supervised pre-training with consistency regularization to bridge the gap between supervised learning and semi-supervised learning during fine-tuning.

\section{Method}
\subsection{Network Architecture}
The network structure of SIAVC is shown in Fig \ref{OverView}. We define labeled data and unlabeled data as:
\begin{equation}
    \begin{gathered}
{\mathcal{X}} = \left\{ {\left( {{x_i},{y_i}} \right),i \in \left( {1,...,I} \right)} \right\}, \\
{\mathcal{U}} = \left\{ {{u_j},j \in (1,...,J)} \right\},
    \end{gathered}
\end{equation}
where ${x_i}\in{\mathbb{R}^{C \times T \times W \times H}}$ represents labeled video samples, $y \in \left\{ {0,1, \cdots ,C} \right\}$ is the classification label for these samples, and ${u_j} \in {\mathbb{R}^{C \times T \times W \times H}}$ represents unlabeled video samples. For each ${u_j}$, it's pseudo-labels ${y'_j} \in \left\{ {0,1, \cdots ,C} \right\}$ will be generated based on the model's previous iterations' results. Additionally, ${\Phi _{emb}}(\cdot)$ is cube embedding, $Enc(\cdot)$ represents the encoder, and $Z$ denotes the latent representation.

\begin{figure}[!t]
  \centering
  \includegraphics[width=0.5\textwidth]{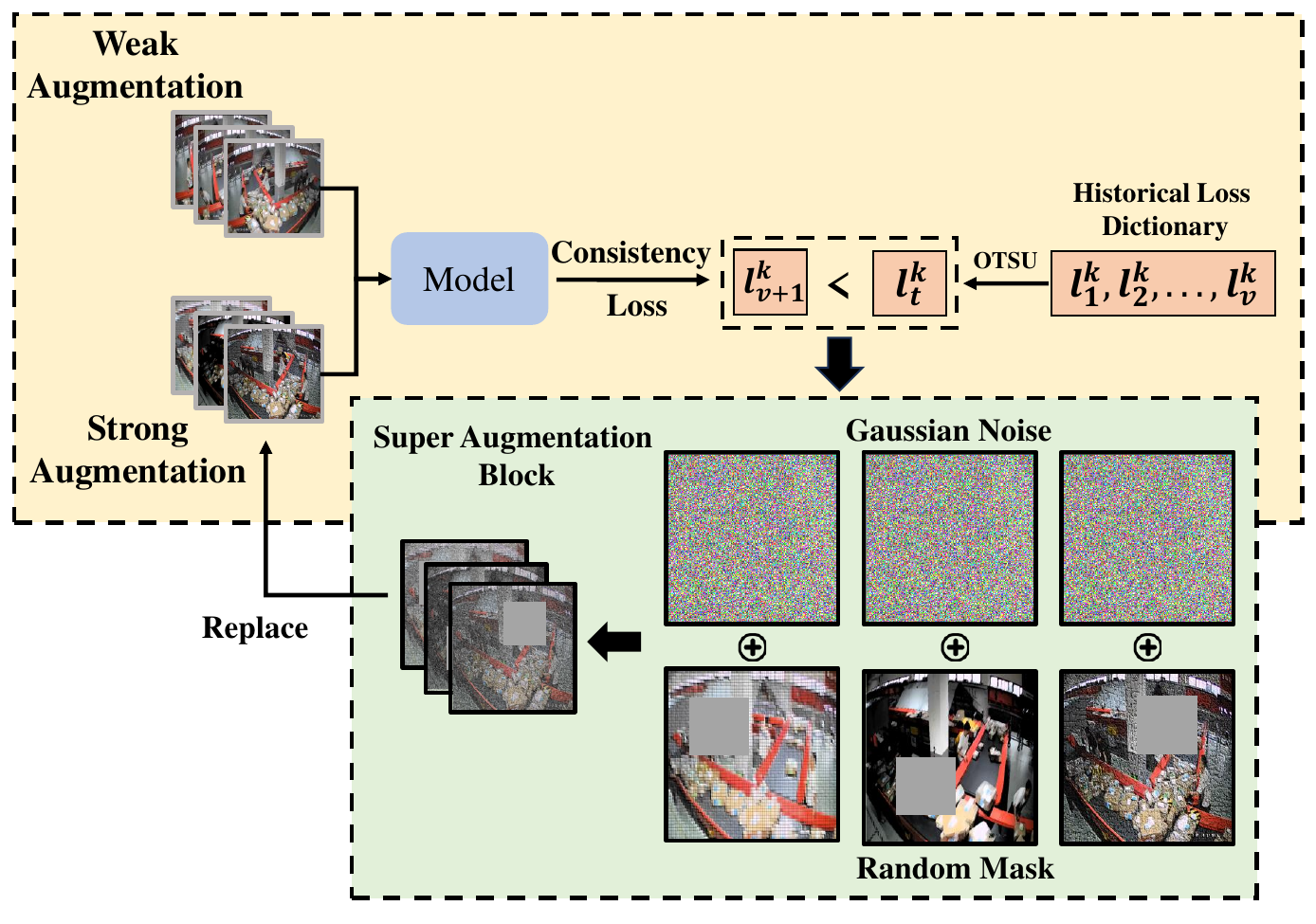}
  \caption{SAB adds Gaussian noise and applies random masking to strongly augmented samples when the consistency loss is lower than a threshold computed based on historical losses.}
  \label{SAB}
\end{figure}

For labeled samples ${x_i}$, we obtain embedding cubes ${\Phi _{emb}}({x_i})$. Then, we send these cubes to encoder $Enc$ to extract the feature $Z$ and train a classifier $Cls$. 
For unlabeled data ${u_j}$, we employ strong augmentation ${\mathcal{A}}({u_j})$ and weak augmentation $\alpha ({u_j})$. After cube embedding, the encoder extracts features, and the classifier $Cls$ outputs predictions and calculates consistency loss ${{\mathcal{L}}_{cons}}$. When ${ {\mathcal{L}}_{cons}}$ falls below a certain threshold, strongly augmented samples will undergo further augmentation by SAB. Then, to fully leverage both labeled and unlabeled samples, VCAM maintains queues ${Q_L}$ and ${Q_U}$ to store labeled and unlabeled samples, respectively. Additionally, it incorporates high-confidence predictions from the model on unlabeled samples into the labeled samples in ${Q_L}$. By randomly selecting samples from these two queues and performing interpolation, it generates more diverse augmented data to augment the training dataset. After cube embedding, the encoder extracts features, and classification is performed by the classifier $Cls$, while the discriminator $Dis$ is responsible for minimizing the distribution gap.

In conclusion, SIAVC is proficient at achieving effective semi-supervised video accident classification. Subsequently, we will provide a comprehensive exposition on SAB in conjunction with consistency regularization and introduce the VCAM in association with adversarial distribution alignment.

\subsection{Super Augmentation Block}
One of the core concepts of consistency regularization is to introduce perturbations to the data and encourage the model to produce consistent output distributions \cite{BD19}. In essence, we apply a flip to unlabeled sample ${u_j}$, which represent as:
\begin{equation}
    \begin{gathered}
    {u_j} = \left\{ {u{x_t}{\text{ ,}}t \in \left( {1,...,T} \right)} \right\},\\
    \end{gathered}
\end{equation} 
where $u{x_t}$ represents video frames, and $T$ is number of frames. Thus, weakly augmented of unlabeled samples are defined as:
\begin{equation}
    \begin{gathered}
    w{u_j} = \left\{ {\alpha (u{x_t}{\text{) ,}} t \in \left( {1,...,T} \right)} \right\}.\\
    \end{gathered}
\end{equation} 
For strong augmentation, we employ RandAugment \cite{CED20} to video frames, and strongly augmented samples are defined as:
\begin{equation}
    \begin{gathered}
    s{u_j} = \left\{ {{\mathcal{A}}(u{x_t}{\text{) ,}}t \in \left( {1,...,T} \right)} \right\}.\\
    \end{gathered}
\end{equation} 
The augmentation techniques involved in RandAugment are listed in Table \ref{Randaugment}.
Then, we computes a consistency loss based on the predictions for $s{u_j}$ and $w{u_j}$.

\begin{table}[!t]
\centering
\caption{Augmentation techniques in Randaugment.}
\resizebox{0.48\textwidth}{!}{
\begin{tabular}{|c|l|}
\hline
Augment type      & Description                                       \\ \hline
Identity     & No change to the original image                   \\ \hline
AutoContrast & Automatically adjusts image contrast              \\ \hline
Equalize     & Equalizes the image histogram                     \\ \hline
Rotate       & Rotates the image                                 \\ \hline
Solarize     & Inverts some pixel values                         \\ \hline
Color        & Adjusts the color balance of the image            \\ \hline
Posterize    & Reduces the number of bits for each color channel \\ \hline
Contrast     & Adjusts the image contrast                        \\ \hline
Brightness   & Adjusts the image brightness                      \\ \hline
Sharpness    & Enhances the sharpness of the image               \\ \hline
ShearX       & Shears the image along the horizontal axis        \\ \hline
ShearY       & Shears the image along the vertical axis          \\ \hline
TranslateX   & Translates the image horizontally                 \\ \hline
TranslateY   & Translates the image vertically                   \\ \hline
\end{tabular}}
\label{Randaugment}
\end{table}

However, in image-based consistency regularization, some strongly augmented samples can still be predicted with high-confidence, even after substantial perturbations have been applied, resulting in diminishing benefits for model optimization with such samples \cite{GEE23}. Given that video contains temporal information in the inter-frame relationships, 3D convolutional neural networks extract image frame features and consider inter-frame temporal features, leading to predict result with high confidence in strongly augmented samples. Therefore, inspired by \cite{GEE23}, we propose a novel re-augmentation module Super Augmentation Block.

The specific details of SAB are illustrated in Fig. \ref{SAB}. During each iteration, the model dynamically maintains a consistency loss dictionary, denoted as:
\begin{equation}
    \begin{gathered}
    Dict = \left\{ {l_1^k,l_2^k,...,l_v^k} \right\}\\
    \end{gathered}
\end{equation}
where $k \in \left\{ {1,2,...,K} \right\}$ represents the index of unlabeled samples, and $v \in \left\{ {1,2,...,V} \right\}$ corresponds to the current iteration count. SAB utilizes OTSU \cite{ONA79} to calculate a loss threshold, denoted as $l_t^k$, for each iteration. Only if $l_{v + 1}^k$ is lower than $l_t^k$, strongly augmented samples are further augment as:
\begin{equation}
    \begin{gathered}
    s{u_j} = \left\{ {\mathcal{M}(\mathcal{G}(\mathcal{A}(u{x_t}{\text{))) ,}} t \in \left( {1,...,T} \right)} \right\},\\
    \end{gathered}
\end{equation}
where $\mathcal{G}(\cdot)$ represents Gaussian noise, and $\mathcal{M}(\cdot)$ signifies random masking. Through this approach, SAB is capable of re-augmenting well-learned augmented samples, thus enabling the model to benefit from consistency regularization.

\textbf{Why Gaussian noise and random masking?} In the real world, intelligent monitoring devices at industrial production sites often exhibit poor performance, and sensors themselves introduce noise during the process of capturing digital images. This noise may arise from factors such as prolonged operation of monitoring devices leading to thermal noise, as well as noise from photoelectric converters. Gaussian noise effectively simulates the noise present in surveillance. Additionally, from a surveillance perspective, some accidents may be briefly obscured by autonomous vehicles or workers. By randomly masking positions in image frames, we can effectively simulate situations where target accidents are momentarily obscured in surveillance videos. In the SAB block, our aim is for the model to learn feature representations of different accidents through more challenging samples, which helps the model adapt to complex scenarios. 

\begin{figure*}
  \centering
  \includegraphics[width=0.9\textwidth]{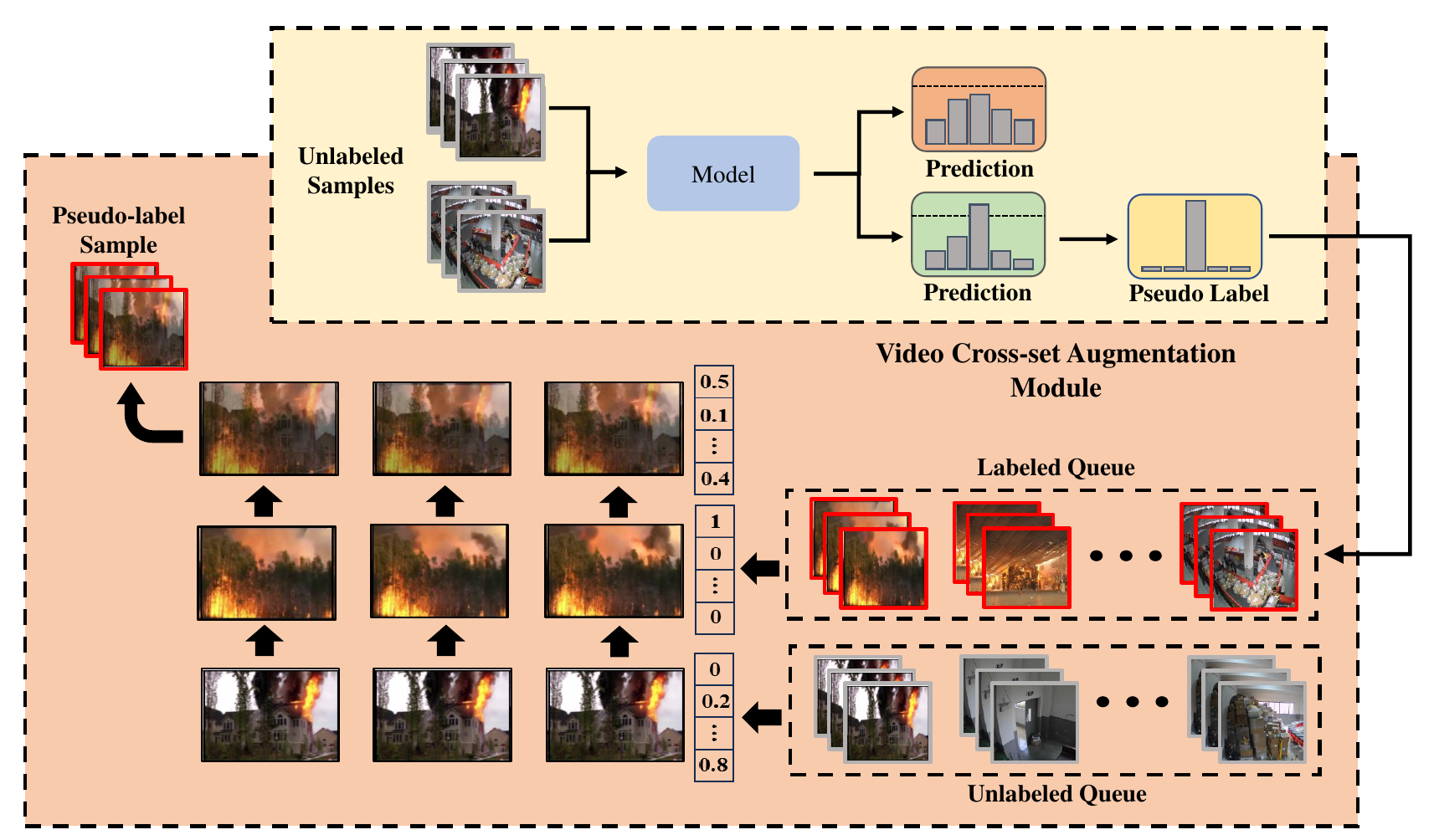}
  \caption{VCAM uses high-confidence unlabeled samples as labeled samples for interpolation in videos, generating more diverse pseudo-label samples.}
  \label{VCAM}
\end{figure*}

\subsection{Video Cross-set Augmentation Module}
In semi-supervised learning, the significant disparity in the quantities of labeled and unlabeled data leads to a sampling bias issue \cite{WQ19}. In other words, the empirical distribution of labeled data deviates from its true distribution.
Adversarial distribution alignment aims to mitigate the negative impact of limited sampling of labeled data on model optimization \cite{WFR22}. In FireMatch \cite{QL23}, we fully use both labeled and unlabeled data by generating pseudo-label samples through interpolation, which we call "video cross-set augmentation samples." Specifically, we multiply the temporal dimension $T$ and the channel dimension $C$ of data samples, resulting in $x_i^{inter}{\text{ }} \in {\mathbb{R}^{E \times W \times H}}$ and $u_j^{inter} \in {\mathbb{R}^{E \times W \times H}}$, respectively, where $E = T{}\cdot{}C$ signifies the magnitude of pixel-wise variations. The interpolated samples can be expressed as: 

\begin{equation}
\label{EQ.1}
    \begin{gathered}
  {\tilde x_i} = {\lambda _{inter}} \cdot {\text{ }}x_i^{inter} + (1 - {\lambda _{inter}}) \cdot {\text{ }}u_j^{inter}, \\
  {\tilde y_i} = {\lambda _{inter}}{\text{ }}\cdot{\text{ }}{y_i} + (1 - {\lambda _{inter}}){\text{ }}\cdot{\text{ }}{y'_j}, \\
  {\tilde z_i} = {\lambda _{inter}}{\text{ }}\cdot{\text{ }}0 + (1 - {\lambda _{inter}}){\text{ }}\cdot{\text{ }}1, \\ 
\end{gathered} 
\end{equation}  
where 0 and 1 represent the discriminator labels for labeled sample and unlabeled sample respectively. ${\lambda _{inter}}$ is a random variable. Additionally, ${\tilde x_i}$ and ${\tilde y_i}$ represent the interpolated augmented samples and their corresponding class labels, with corresponding discriminator labels denoted as ${\tilde z_i}$.

However, generated pseudo-label samples face the challenge of limited diversity when there are insufficient labeled samples. As illustrated in Fig. \ref{VCAM}, we introduce VCAM to incorporate high-confidence unlabeled samples generated by the model into the generation queue as pseudo-label samples. VCAM aims to produce more diverse set of pseudo-label samples, alleviating the impact of sampling experience mismatch and expand training data. Specifically, for each unlabeled sample ${u_j}$, the model's predicted result is ${P_c} = \{ {p_1},{p_2},..{p_C}\} $. Unlabeled sample is designated as $({x_j},{y_j} = c)$ when $\max({P_c})$ is grater than $\tau $. Then, the labeled and unlabeled queue for interpolation are as follows:

\begin{equation}
\label{EQ.2}
    \begin{gathered}
{Q_L} = \{ ({x_i},{y_i}),({x_j},{y_j})|i \in (1,...,I),j \in (1,...,J)\},  \\
{Q_U} = \{ {u_j}|j \in (1,...,J)\}.\\
\end{gathered} 
\end{equation}
In each iteration, the order of ${Q_U}$ and ${Q_L}$ is randomly shuffled, and diverse video cross-set augmentation samples are generated using the interpolation method outlined in Eq. \ref{EQ.1}.

\begin{figure*}[!t]
  \centering
  \includegraphics[width=0.9\textwidth]{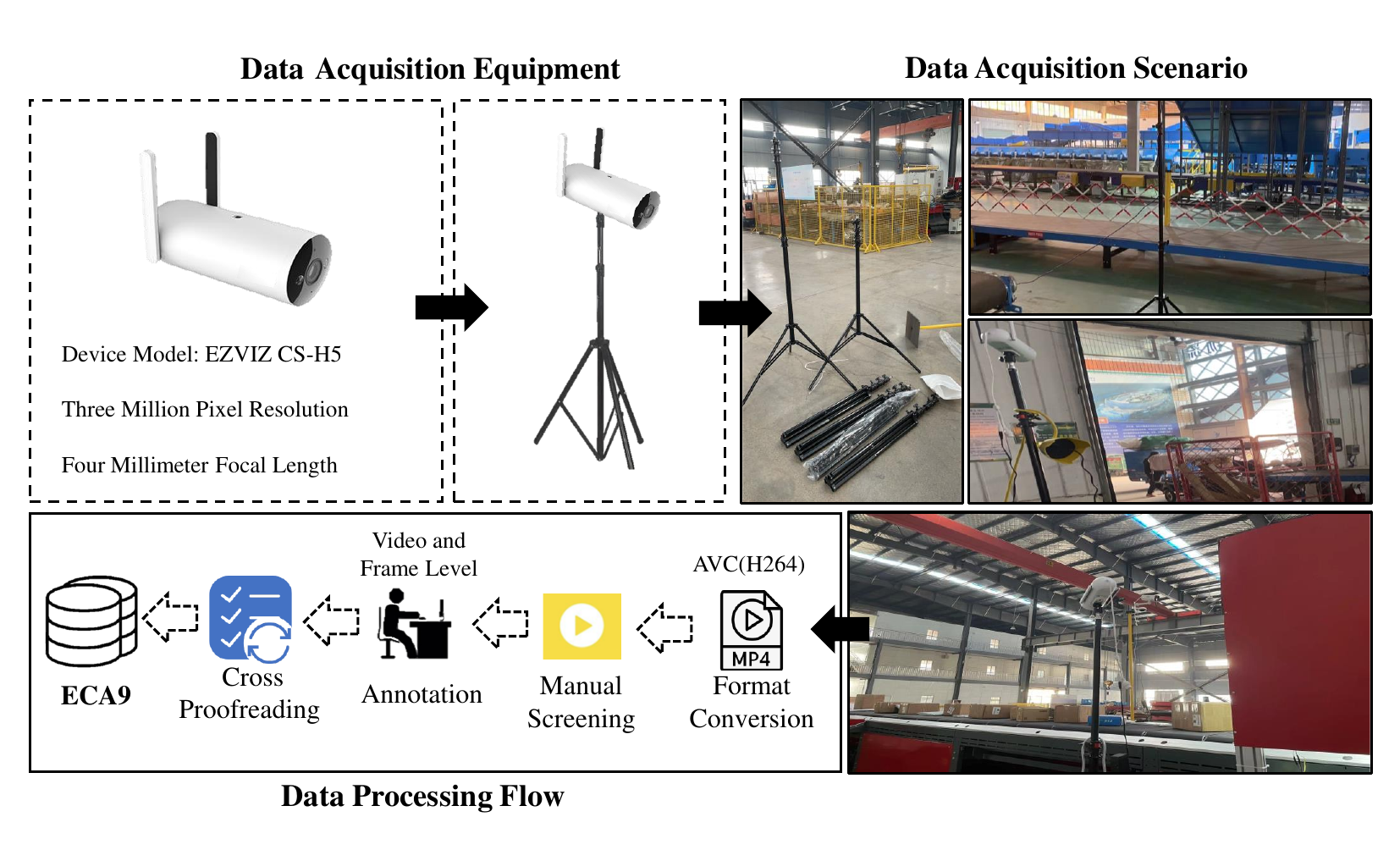}
  \caption{Express Center Accidents 9 dataset construction process.}
  \label{BD1}
\end{figure*}

\subsection{Loss Function}
The overall loss of the proposed method consists of four components: a classifier's supervised loss for labeled samples ${{\mathcal L}}_{cls}$, a adversarial distribution alignment loss for unlabeled data ${{\mathcal L}}_{align}$, a consistency loss ${{\mathcal L}}_{cons}$, and a class fairness loss ${{\mathcal L}}_{fair}$, which drives balanced predictions in the model. Let $Cls\left( {y|{x_i}} \right)$ represent the probability that the classifier predicts for input ${x_i}$. The definition of the supervised loss is as follows:

\begin{equation}
\label{EQ.3}
    \begin{gathered}
{{\mathcal{L}}_{cs}} = \frac{1}{I}\sum\limits_{i = 1}^I {{\mathcal{H}}({y_i},Cl{s_{cs}}(y|{x_i}))},\\
\end{gathered} 
\end{equation}
where ${\mathcal{H}}(\cdot)$ represents the cross-entropy loss function. As for the unsupervised distribution alignment loss, it is defined as:

\begin{equation}
\label{EQ.4}
    \begin{aligned}
        {\mathcal{L}}_{\text{align}} &= \frac{1}{I}\sum_{i=1}^{I} \rho \cdot \mathcal{H}\left( {\tilde{y}_i, \text{Cls}(y|\tilde{x}_i)} \right) + \\
        &\quad \lambda_{\text{inter}} \cdot \mathcal{H}\left( {\tilde{z}_i, \text{Dis}(z|\tilde{x}_i)} \right),
    \end{aligned}
\end{equation}
where ${\tilde x_i}$ stands for pseudo-label sample generated by VCAM, ${\tilde y_i}$ represents it's label, ${\tilde z_i}$ denotes the discriminator labels, and $Dis(z|{\tilde x_i})$ signifies the discriminator's output results. Inspired by FreeMatch \cite{WY22}, we define the consistency loss as:

\begin{equation}
\label{EQ.5}
    \begin{gathered}
{ {\mathcal{L}}_{cons}} = \frac{1}{J}\sum\limits_{j = 1}^J {1\left( {\max \left( {{q_j}} \right) > {\tau _{final}}(\arg \max ({q_j}))} \right)} {\text{ }} \\ \cdot {\text{ }} {\mathcal{H}}({\hat q_j},{Q_j}), \\
\end{gathered} 
\end{equation}
where ${Q_j} = Cls\left( {y|s{u_j}} \right)$ and ${q_j} = Cls\left( {y|w{u_j}} \right)$ represent the classifier's predicted probabilities for strong and weak augmented samples, respectively, and ${\hat q_j}$ is the ``one-hot" label of ${q_j}$. Additionally, ${\tau _{final}} = MaxNorm({\tilde p_j}\left( n \right)){\text{ }}{}\cdot{\tau _{global}}$ is the pseudo-label threshold for unlabeled samples, calculated by FreeMatch through the computation of local thresholds ${\tilde p_j}$ for each class and the global threshold ${\tau _{global}}$. At the same time, we introduce the class fairness loss:

\begin{equation}
\label{EQ.6}
    \begin{gathered}
{ {\mathcal{L}}_{fair}} =  - {\mathcal{H}}(SumNorm(\frac{{{{\tilde p}_j}}}{{{{\tilde h}_j}}}),SumNorm(\frac{{\bar p}}{{\bar h}})), 
\end{gathered} 
\end{equation}
where ${\tilde h_j}$ is the histogram distribution of local thresholds. similarly, $\bar p$ represents the current sample's prediction results, and $\bar h$ is the corresponding histogram distribution, with $SumNorm = ( \cdot )/\sum {( \cdot )} $. Optimizing the fairness objective encourages the reduction of bias toward individual classes. Finally, we define the total loss as:

\begin{equation}
\label{EQ.7}
    \begin{gathered}
{\mathcal{L}} = {{\mathcal{L}}_{cs}} + {\omega _a}{{\mathcal{L}}_{align}} + {\omega _c}{{\mathcal{L}}_{cons}} + {\omega _f}{{\mathcal{L}}_{fair}},
\end{gathered} 
\end{equation}
where ${\omega _a}$, ${\omega _c}$, and ${\omega _f}$ are the weight parameters for the adversarial distribution alignment loss, the consistency loss, and the class fairness loss, respectively.

\begin{figure*}[!t]
  \centering
  \includegraphics[width=\textwidth]{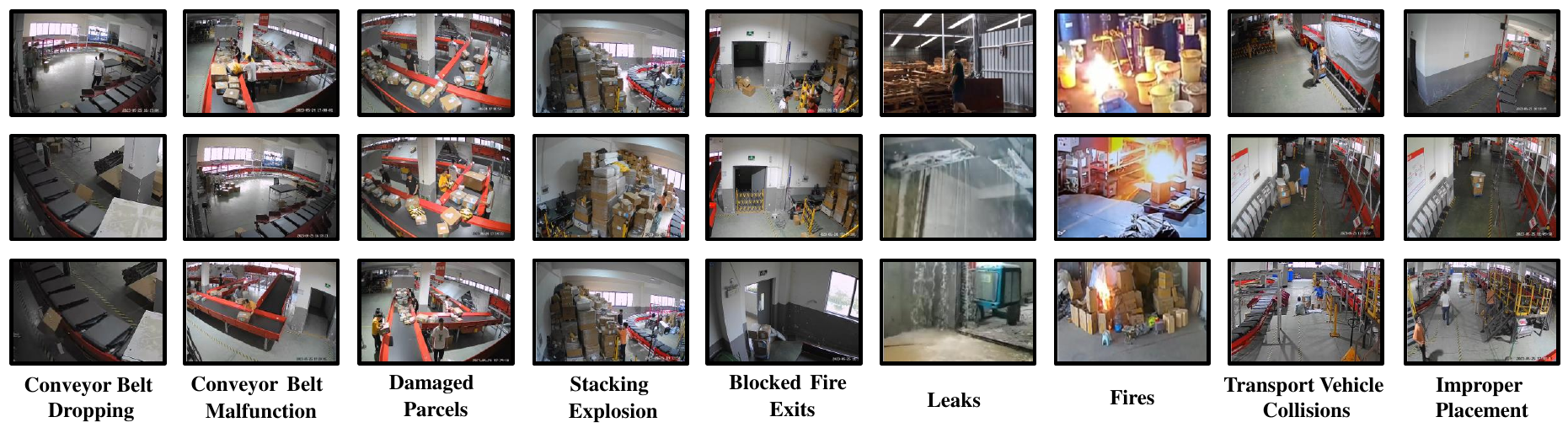}
  \caption{Video examples for each category of Express Center Accidents 9.}
  \label{ECA9}
\end{figure*}

\section{Express Center Accidents 9}

The construction process of the ECA9 dataset is shown in Fig. \ref{BD1}. We use eight surveillance cameras of the EZVIZ CS-H5 model with support rods. The camera has a focal length of four millimeters and has three megapixels, and the video resolution is 1920×1080. We deploy these cameras in the hub-level express sorting sites for long-term data collection.

After completing the data collection, we first converted the collected video data format to AVC (H264), and then used the video player and FFmpeg to screening and cropping video. Secondly, we classify the preprocessed videos and annotate them frame by frame. Finally, we cross- proogread the annotated data to construct the ECA9 dataset.

As shown in Fig. \ref{ECA9}, ECA9 contains nine classes of typical accidents in hub-level express processing centers, including Conveyor Belt Dropping, Conveyor Belt Malfunction, Damaged Parcels, Stacking Explosion, Blocked Fire Exits, Leaks, Fires, Transport Vehicle Collisions, and Improper Placement. Due to the uncontrollability of fire and the risk of equipment damage and electric shock caused by leaks, we only simulate a limited amount of data with the help of engineers. The quantity of each category in the dataset and the number of frames are detailed in Table \ref{t_dataset}.

To the best of our knowledge, ECA9 is the first accident dataset for hub-level express processing centers. We not only provide video-level labels for classification tasks, but we also provide frame-level anomaly labels for anomaly detection tasks. In the future, we will open more and more reliable industrial accident data to promote the development of this field.

\begin{table}[!h]
\renewcommand{\arraystretch}{1.1}
\centering
\caption{The details of the ECA9 dataset.}
\resizebox{0.4\textwidth}{!}{
\begin{tabular}{ccc}
\hline
\multicolumn{3}{c}{Express Center Accidents 9}                                                     \\ \hline
\multicolumn{1}{c|}{Class}                        & \multicolumn{1}{c|}{Num} & Frames                     \\ \hline
\multicolumn{1}{c|}{Conveyor Belt Dropping}       & \multicolumn{1}{c|}{14}  & 2,086                      \\
\multicolumn{1}{c|}{Conveyor Belt Malfunction}    & \multicolumn{1}{c|}{18}  & 2,492                      \\
\multicolumn{1}{c|}{Damaged Parcels}              & \multicolumn{1}{c|}{136} & 20,196                     \\
\multicolumn{1}{c|}{Stacking Explosion}           & \multicolumn{1}{c|}{14}  & 2,086                      \\
\multicolumn{1}{c|}{Blocked Fire Exits}           & \multicolumn{1}{c|}{59}  & 8,791                      \\
\multicolumn{1}{c|}{Leaks}                        & \multicolumn{1}{c|}{36}  & 3,306                      \\
\multicolumn{1}{c|}{Fires}                        & \multicolumn{1}{c|}{45}  & 5,524                      \\
\multicolumn{1}{c|}{Transport Vehicle Collisions} & \multicolumn{1}{c|}{37}  & 5,513                      \\
\multicolumn{1}{c|}{Improper Placement}           & \multicolumn{1}{c|}{87}  & 12,963                     \\ \hline
\multicolumn{1}{c|}{Total}                        & \multicolumn{1}{c|}{446} & 62,957                     \\ \hline
\end{tabular}
}
\label{t_dataset}
\end{table}

\begin{figure}[!t]
  \centering
  \includegraphics[width=0.4\textwidth]{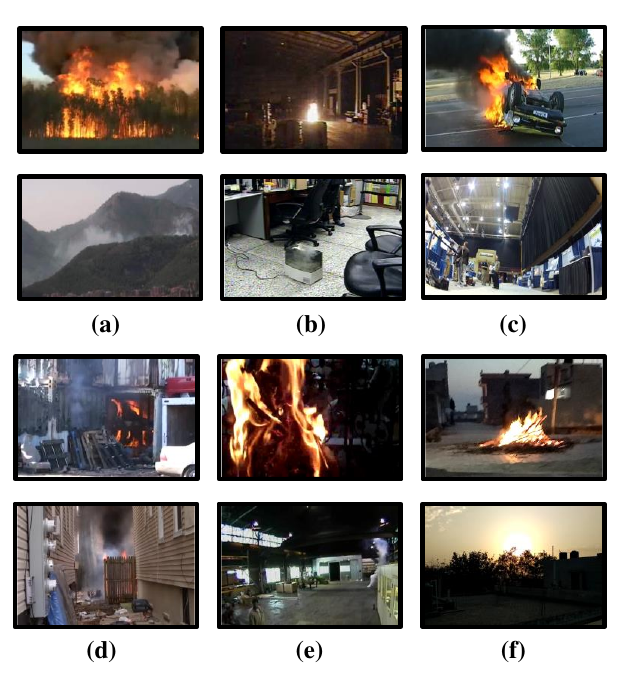}
  \caption{Examples of publicly available fire-related video datasets. (a) MIVIA Fire and Smoke. (b) KMU Fire and Smoke Database. (c) Furg Fire Dataset. (d) LAD2000. (e) Firesense. (f) Custom-Compiled Fire Dataset.}
  \label{Fire}
\end{figure}

\begin{table}[!h]
\renewcommand{\arraystretch}{1.2}
\centering
\caption{The details of the Fire Detection dataset.}
\resizebox{0.45\textwidth}{!}{
\begin{tabular}{ccc}
\hline
\multicolumn{3}{c}{Fire Detection}                                                                        \\ \hline
\multicolumn{1}{c|}{Dataset Name}                              & \multicolumn{1}{c|}{Fire} & Non-Fire                  \\ \hline
\multicolumn{1}{c|}{MIVIA Fire and Smoke \cite{DLR14,FP15}}         & \multicolumn{1}{c|}{14}  & 166                      \\
\multicolumn{1}{c|}{KMU Fire and Smoke Database \cite{KBC11}}  & \multicolumn{1}{c|}{26}  & 12                      \\
\multicolumn{1}{c|}{Furg Fire Dataset \cite{HV17}}                         & \multicolumn{1}{c|}{17} & 6                     \\
\multicolumn{1}{c|}{LAD2000 \cite{WB21}}                                   & \multicolumn{1}{c|}{107}  & -                      \\
\multicolumn{1}{c|}{Firesense \cite{KD15}}                                 & \multicolumn{1}{c|}{17}  & 32                      \\
\multicolumn{1}{c|}{Custom-Compiled Fire Dataset \cite{JA19}}              & \multicolumn{1}{c|}{45}  & 16                      \\ \hline
\multicolumn{1}{c|}{Total}                                     & \multicolumn{1}{c|}{226} & 232                     \\ \hline
\end{tabular}
}
\label{t_dataset2}
\end{table}

\section{Experiment}
\subsection{Implementation Details}
At the training stage, we set the initial learning rate to $\eta  = 0.03$ and use cosine learning rate decay \cite{IL16} to update it to $\eta \cos \left( {\frac{{7\pi {g}}}{{16G}}} \right)$, where ${g}$ denotes the current iteration count, $G$ is total number of training steps. We use SGD \cite{SI13} with a momentum of 0.9 and set weight decay to 0.0005. Inspired by study \cite{TZV22}, we use VisionTransformer \cite{DAA20} as the backbone for all methods to ensure fair comparison. Additionally, we implement all methods in Python 3.9 with PyTorch 2.0.1 and use a 12-core computer equipped with an Intel Xeon Silver 4214R 2.4 GHz (128GB RAM) CPU and seven NVIDIA GeForce RTX 3090 24G GPUs for training and testing.


\begin{table*}[t!]
\caption{Comparison with state-of-the-art semi-supervised classification methods on ECA9 and Fire Detection datasets. Top-1 and Top-5 accuracy (\%) are used as evaluation metrics.}
\renewcommand{\arraystretch}{1.2}
\resizebox{\textwidth}{!}{
\begin{tabular}{cclclclcclclcll}
\cline{1-14}
Dataset              & \multicolumn{6}{c}{Express Center Accidents 9}                                                                                                                 &                      & \multicolumn{6}{c}{Fire Detection}                                                             &  \\ \cline{2-7} \cline{9-14}
Method               & \multicolumn{2}{c}{134 labels}  & \multicolumn{2}{c}{80 labels}   & \multicolumn{2}{c}{28 labels}   & & \multicolumn{2}{c}{134 labels}          & \multicolumn{2}{c}{80 labels}           & \multicolumn{2}{c}{28 labels}      &  \\ 
\cline{1-14}
                     & Top-1          & Top-5          & Top-1          & Top-5          & Top-1          & Top-5 &          & \multicolumn{2}{c}{Top-1}               & \multicolumn{2}{c}{Top-1}               & \multicolumn{2}{c}{Top-1}          & \\
Temporal Ensembling  & 81.46          & 98.88          & 64.04          & 93.46          & 64.04          & 92.13 &          & \multicolumn{2}{c}{84.24}               & \multicolumn{2}{c}{84.24}               & \multicolumn{2}{c}{76.09}          & \\
Mean Teacher         & 78.09          & 98.31          & 74.16          & 97.75          & 50.00          & 93.26 &          & \multicolumn{2}{c}{86.41}               & \multicolumn{2}{c}{85.33}               & \multicolumn{2}{c}{82.07}          & \\
FixMatch             & 82.02          & 96.07          & 76.92          & 94.94          & 67.42          & 91.57 &          & \multicolumn{2}{c}{85.87}               & \multicolumn{2}{c}{84.24}               & \multicolumn{2}{c}{79.89}               & \\
UPS                  & 73.59          & 98.88          & 73.03          & 89.32          & 44.94          & 87.64 &          & \multicolumn{2}{c}{83.69}               & \multicolumn{2}{c}{83.15}               & \multicolumn{2}{c}{82.07}          & \\
FlexMatch            & 83.15          & 97.75          & 71.91          & 98.31          & 59.55          & 84.83 &          & \multicolumn{2}{c}{86.69}               & \multicolumn{2}{c}{\textbf{86.41}}      & \multicolumn{2}{c}{84.24}               & \\
AdaNet               & 80.90          & 97.75          & 78.09          & 91.57          & 62.92          & 92.13 &          & \multicolumn{2}{c}{86.41}               & \multicolumn{2}{c}{83.15}               & \multicolumn{2}{c}{81.52}               &      \\
FreeMatch            & 85.39          & 96.63          & 82.58          & \textbf{98.31} & 62.36          & 89.33 &          & \multicolumn{2}{c}{85.87}               & \multicolumn{2}{c}{82.07}               & \multicolumn{2}{c}{78.26}          & \\
FireMatch            & 81.46          & 98.88          & 71.91          & 91.57          & 50.56          & 83.37 &          & \multicolumn{2}{c}{87.50}               & \multicolumn{2}{c}{85.33}               & \multicolumn{2}{c}{82.61}          & \\
SIAVC          & \textbf{88.76} & \textbf{99.44} & \textbf{87.08} & 97.19          & \textbf{68.54} & \textbf{94.38} & & \multicolumn{2}{c}{\textbf{89.13}}      & \multicolumn{2}{c}{85.33}               & \multicolumn{2}{c}{\textbf{84.24}} & \\ 
\cline{1-14} 
\end{tabular}
}
\label{t_result}
\end{table*}

\subsection{Datasets}
Besides the ECA9 dataset, we collect and integrate publicly available video datasets for fire detection, denoted as Fire Detection. The composition of Fire Detection is shown in Table \ref{t_dataset2}, which includes datasets from \cite{DLR14,FP15,KBC11,HV17,WB21,KD15,JA19}.
As shown in Fig. \ref{Fire}, Fire Detection comprises 226 videos featuring evident fire characteristics and 232 videos containing smoke, moving light sources, or moving red objects.
The video data are uniformly set to a resolution 320×240 and a frame rate of 15 FPS. We divide these datasets into a 6:4 split for training and testing to facilitate model performance evaluation.

\subsection{Baseline Methods}
We adapt existing semi-supervised classification algorithms to ensure a fair comparison.

\textbf{Temporal Ensembling} \cite{LST16} introduces self-ensembling to make the same predictions for unlabeled data under various regularization and augmentation conditions.

\textbf{Mean Teacher} \cite{TA17} employs a "teacher" network to ensure consistency with a "student" network, thus promoting similar predictions for labeled and unlabeled data.

\textbf{FixMatch} \cite{SK20} combines consistency regularization with pseudo-labeling, setting pseudo-labels for unlabeled samples with confidence exceeding 0.95 to participate in training.

\textbf{UPS} \cite{RMN21} enhances pseudo-labels' accuracy by significantly reducing the amount of noise encountered during the training process.

\textbf{FlexMatch} \cite{ZB21} dynamically adjusts pseudo-label threshold based on the learning status of samples as determined by current model.

\textbf{AdaNet} \cite{WF22} minimizes the training error of labeled data and the empirical distribution gap between labeled and unlabeled data, effectively constraining the generalization error in semi-supervised learning.

\textbf{FreeMatch} \cite{WY22} improves FixMatch by introducing a dynamic threshold adjustment strategy to obtain more pseudo-label samples, accelerating convergence.

\textbf{FireMatch} \cite{QL23} introduces video cross-set augmentation for generating pseudo-label samples and alleviates the problem of sampling empirical mismatch.

\subsection{Experiment Results and Analysis}
We conduct extensive experiments on Fire Detection and ECA9 datasets. Three different numbers of labels (134 labels ($\approx$50\%), 80 labels ($\approx$30\%), and 28 labels ($\approx$10\%)) are employed to evaluate the performance of various semi-supervised classification algorithms, with each class having labels allocated proportionally to its total size. Top-1 and Top-5 accuracy are utilized as evaluation criteria. 

\begin{figure}[!t]
  \centering
  \includegraphics[width=0.5\textwidth]{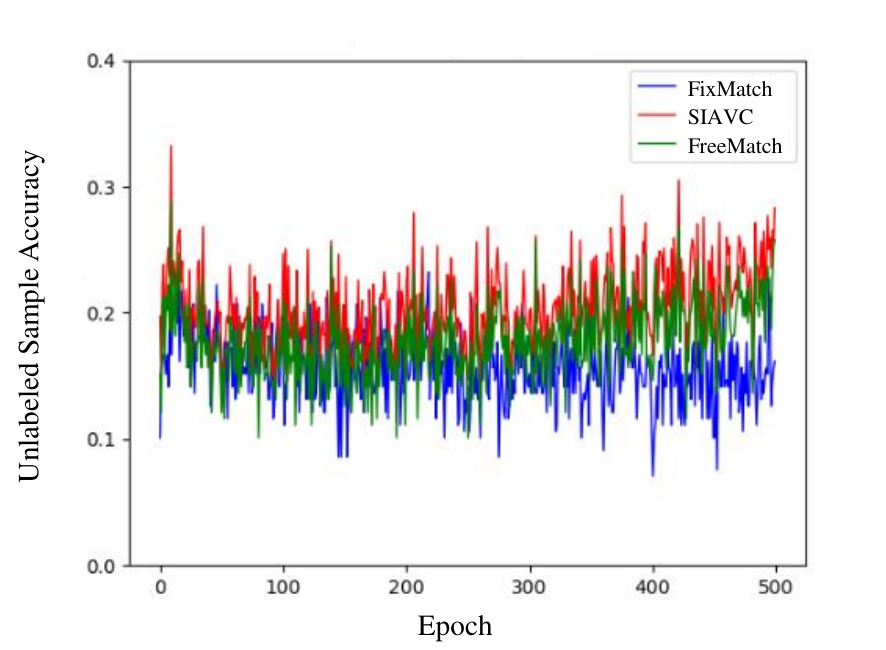}
  \caption{Accuracy of FixMatch, FreeMatch and SIAVC in predicting pseudo-labels for unlabeled samples when only 28 labels ($\approx$10\%) of the labeled samples are used for training.}
  \label{unlabeled_acc}
\end{figure}

Comparison experiment results are shown in Table \ref{t_result}. SIAVC achieves promising results on both datasets with three different label settings. It is worth noting that on ECA9, when the number of labels decreases from 134 labels to 80 labels, the Top-1 accuracy of SIAVC only decreases from 88.76\% to 87.08\%, while the Top-5 accuracy drops by 2.25\%. In contrast, AdaNet and FreeMatch experience a 2.81\% decrease in Top-1 accuracy. FlexMatch and FixMatch, constrained by the number of label samples, see a reduction of 11.24\% and 5.1\% in Top-1 accuracy, respectively. When the number of labels decreases, the limited number of labeled samples cannot support accurate predictions for unlabeled data on the one hand. On the other hand, it is affected by the differences in sampled empirical distributions, which limit performance.

\begin{figure*}[!t]
  \centering
  \includegraphics[width=0.8\textwidth]{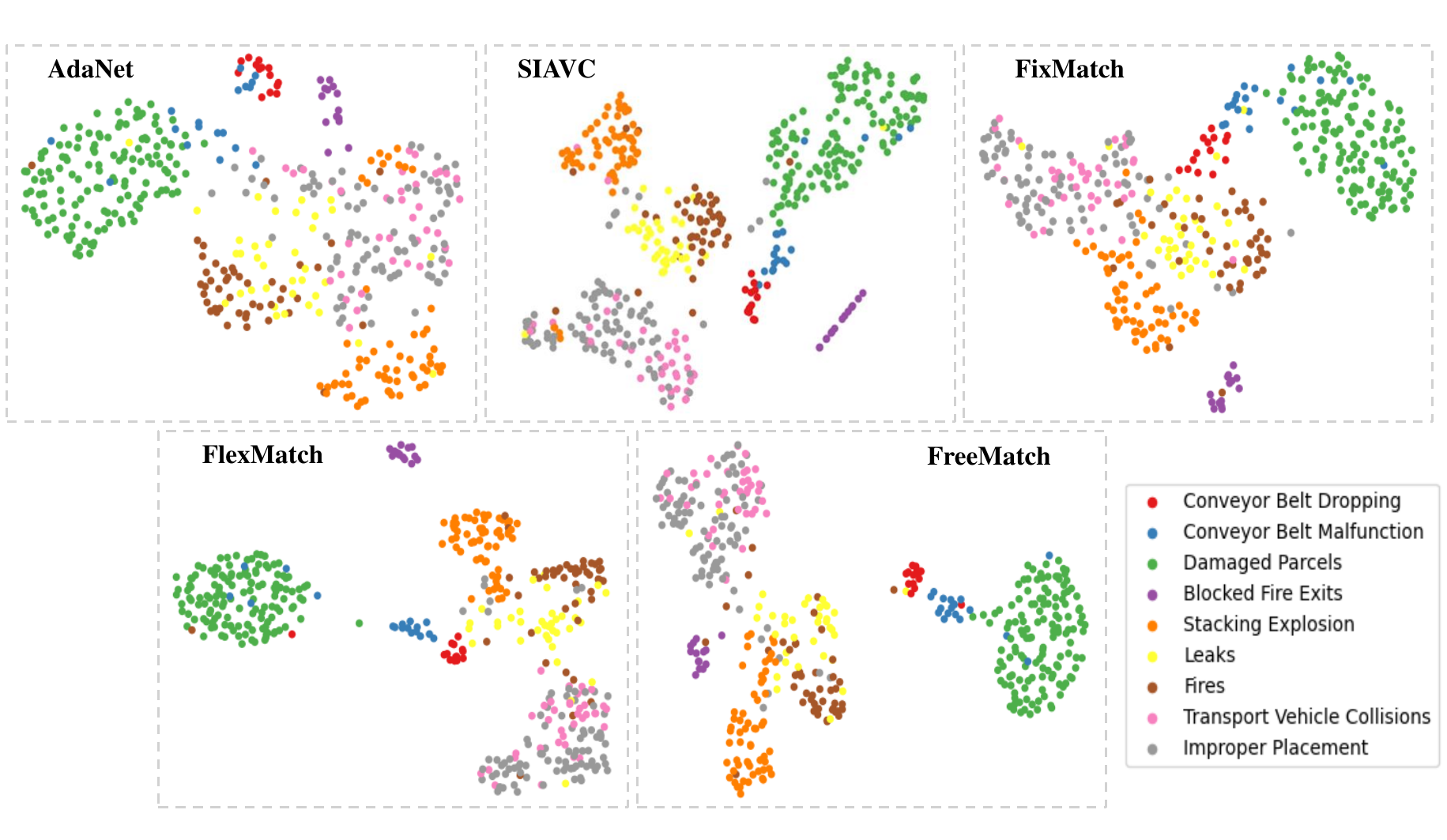}
  \caption{UMAP (sizes=20, n\_neighbors=30, min\_dist=0.6) feature visualization on ECA9 dataset.}
  \label{UMAP}
\end{figure*}

Temporal Ensembling and Mean Teacher, as classic semi-supervised classification algorithms, achieve Top-1 accuracies of 81.46\% and 78.09\%, respectively, on ECA9 with 134 labels. When the number of labels reduces to 80, the impact on Mean Teacher is far less significant compared to Temporal Ensembling. Comparing the results with 28 labels, Temporal Ensembling has a Top-1 accuracy of 64.04\%, with slightly different Top-5 accuracy. When the number of labeled samples is deficient, Temporal Ensembling relies mainly on unlabeled samples for learning, and the number of labeled samples does not dominate the training process.
FireMatch and UPS do not perform well in the nine-class classification task. The main reason is that FireMatch is a semi-supervised model designed for binary fire detection, and its separation of the prediction head from the classifier does not train the classifier's classification ability well for multi-class tasks. UPS relies on a large number of labeled samples for multiple iterations to generate high confidence in unlabeled samples, and insufficient labeled samples for each class make it difficult to predict unlabeled samples correctly.

Additionally, under the 28 labels setting, we conduct statistical analysis on the accuracy of the three methods in predicting unlabeled samples during the training. As shown in Fig. \ref{unlabeled_acc}, in the early stages of training, FixMatch, FreeMatch, and SIAVC all exhibit relatively low accuracy in predicting unlabeled samples. As iterations progress, the accuracy of SIAVC and FreeMatch fluctuates upwards, but FixMatch is constrained in its ability to predict accurately under conditions of limited labeled data due to its higher pseudo-label threshold.

Furthermore, we provide the results of feature space visualization for different methods in Fig. \ref{UMAP}. It's evident that except for Damaged Parcels, AdaNet, and FixMatch have unclear boundaries for the rest of the categories in the ECA9 dataset. The number of Damaged Parcels samples is much higher than other categories, allowing all algorithms to identify them effectively. Additionally, SIAVC, FlexMatch, and FreeMatch achieve good representations. 
However, their ability to distinguish between Transport Vehicle Collisions and Improper Placement could be improved, with FreeMatch and FlexMatch even confusing the two categories in the feature space. In comparison, SIAVC slightly confuses these two types of samples. Transport Vehicle Collisions and Improper Placement scenarios are often similar. They may occur together, posing a challenge to the model's classification capability.

In contrast to accident classification on ECA9, Fire Detection only needs to distinguish between fire and non-fire. Therefore, various algorithms can perform well even in situations with few labels. Particularly, SIAVC achieves a Top-1 accuracy of 89.13\% with 134 labels, thanks to VCAM to generate different pseudo-label fire samples, thereby expanding the training data. It's worth noting that in the 80 labels setting, FlexMatch achieves the best Top-1 accuracy of 86.41\%. The reduction in the number of labels has a small impact. The main reason is that FlexMatch can adjust confidence threshold levels by learning the fire category's state, and the quantity of labeled samples at this point is already sufficient for the model to learn the category's features effectively. A similar situation also occurs with UPS, as the binary classification label quantity is enough to support the model in making many correct predictions for unlabeled samples through extensive iterations. When the number of labeled samples reduces to 28 labels, all the algorithms still manage to achieve good classification accuracy. Both FlexMatch and SIAVC attain a Top-1 accuracy of 84.24\%. FixMatch and FreeMatch also achieve Top-1 accuracies of 79.89\% and 78.26\%, respectively. FixMatch, on the one hand, discards a significant number of correctly predicted pseudo-label samples due to its high confidence threshold. FreeMatch uses a lower threshold in the early stages of training, which leads to a considerable number of incorrect predictions for pseudo-label samples.

\begin{table}[h]
\centering
\caption{The component ablation study of SIAVC and comparative experiments under supervised learning.}
\renewcommand{\arraystretch}{1.2}
\resizebox{0.48\textwidth}{!}{
\begin{tabular}{c|l|cc}
\hline
\multirow{2}{*}{Index} & \multicolumn{1}{c|}{\multirow{2}{*}{Method}} & \multicolumn{2}{c}{ECA9} \\ \cline{3-4} 
& \multicolumn{1}{c|}{} & \multicolumn{1}{c|}{Top-1 (\%)} & Top-5 (\%) \\ \hline
1 & CR+FT & \multicolumn{1}{c|}{82.02} & 96.07 \\
2 & CR+SAT & \multicolumn{1}{c|}{83.71} & 94.38 \\
3 & CR+SAT+FL & \multicolumn{1}{c|}{85.39} & 96.63 \\
4 & CR+SAT+FL+SAB & \multicolumn{1}{c|}{86.51} & 97.75 \\
5 & CR+SAT+FL+SAB+VCAM & \multicolumn{1}{c|}{\textbf{88.76}} & \textbf{99.44} \\ 
\hline
6 & Supervised (with 50\% labels) & \multicolumn{1}{c|}{78.65} & 93.26 \\
7 & Fully-Supervised (with 100\% labels) & \multicolumn{1}{c|}{84.83} & 97.75 \\ \hline
\end{tabular}}
\label{t_as}
\end{table}

\subsection{Ablation Study}
As shown in Table \ref{t_as}, we conduct ablation experiments on ECA9 with 134 labels to validate the effectiveness of various components of SIAVC. Index 1 represents Consistency Regularization (CR) with a fixed pseudo-label threshold of 0.95, Index 2 represents CR with the self-adaptive threshold (SAT) of pseudo-label, and Index 3 introduces Fairness Loss (FL) on top of Index 2. In this setting, we set Index 1-3 as the baseline ablation experiments to assess the effectiveness of combining Consistency Regularization with a dynamic pseudo-label threshold. The results demonstrate that CR combined with SAT and FL significantly improves the performance of semi-supervised classification.

\begin{table}[!h]
\centering
\caption{Evaluation of confidence threshold for unlabeled samples in VCAM.}
\renewcommand{\arraystretch}{1.2}
\resizebox{0.48\textwidth}{!}{
\begin{tabular}{c|c|cc|cc}
\hline
\multirow{2}{*}{Index}       & \multirow{2}{*}{Threshold}       & \multicolumn{2}{c|}{ECA9}                  & \multicolumn{2}{c}{Fire Detection} \\ \cline{3-6} 
                       &                            & Top-1 (\%)          & Top-5 (\%)           & \multicolumn{2}{c}{Top-1 (\%)}       \\ \hline
1                      & 0.4                        & 83.15               & 98.31                & \multicolumn{2}{c}{87.50}               \\
2                      & 0.5                        & 82.58               & 98.88                & \multicolumn{2}{c}{84.24}               \\
3                      & 0.6                        & 82.58               & 98.88                & \multicolumn{2}{c}{\textbf{89.13}}               \\
4                      & 0.7                        & 85.39               & 97.75                & \multicolumn{2}{c}{82.07}               \\
5                      & 0.8                        & 85.39               & 98.88                & \multicolumn{2}{c}{77.72}               \\
6                      & 0.9                        & \textbf{88.76}      & \textbf{99.44}       & \multicolumn{2}{c}{88.04}          \\
7                      & 0.95                       & 85.96               & 97.19                & \multicolumn{2}{c}{84.78}        \\ \hline
\end{tabular}
}
\label{t_vcam}
\end{table}

Furthermore, we also perform ablation experiments on the proposed SAB and VCAM, as shown in Indexes 4 and 5. The experimental results indicate that SAB and VCAM further improve classification accuracy.
Finally, Index 6 and Index 7 represent ablation experiments for the VideoTransformer \cite{NDV21} using only 134 labels and providing all training sample labels, respectively. The experimental results show that the proposed approach, incorporating VCAM, surpasses the performance of fully-supervised VideoTransformer on ECA9 in semi-supervised classification. SIAVC correctly predicts unlabeled samples through SAB and generates diverse pseudo-label samples through VCAM to augment the training data.

In addition, we evaluate the confidence threshold involved in VCAM under the same settings, as shown in Table \ref{t_vcam}. From the experimental results, a threshold setting of 0.9 is optimal for the 9-class classification task. On the other hand, for the 2-class classification task, we recommend adjusting the threshold to 0.6 to achieve the best model performance. Setting the confidence threshold too low can result in erroneous pseudo-label samples in the synthetic queue. Conversely, setting the threshold too high hinders unlabeled samples from entering the synthetic queue, even if they have been correctly predicted. Therefore, the threshold setting for VCAM should be determined based on the task's difficulty level.

\section{Future Work}
Although SIAVC has achieved encouraging results on the provided datasets, it is still far from perfect. In the future, we will discuss the impact of sample gradient updates on SAB and the improvement of completely obscured targets through adaptive mask resizing. Additionally, the threshold currently used by VCAM is fixed, which may limit its ability to provide data beyond the training set distribution for the model. As with many semi-supervised tasks, an excellent dynamic threshold adjustment strategy is also crucial for improving VCAM performance. Furthermore, we will continue to provide more accident video data, including but not limited to scenarios of express processing, in the hope of continuously driving the development of artificial intelligence in industrial safety.

\section{Conclusion}
This paper proposes a semi-supervised video classification framework for accidents called SIAVC. We propose SAB to re-augment the strongly augmented samples that have already been well-learned in consistency regularization, allowing the model optimization to continue benefiting from these strongly augmented samples. We design VCAM to incorporate high-confidence unlabeled samples as labeled samples for pseudo-label generation. VCAM alleviates the problem of monotonic generation of samples with scarce labeled data while mitigating sampling experience mismatch between labeled and unlabeled data. 
Moreover, we construct a hub-level express center accident dataset, including video-level labels and frame-level anomaly annotations. In future work, we will further compress the model size through knowledge distillation and continue to collect reliable accident video data.

\bibliographystyle{IEEEtran}
\bibliography{IEEEabrv,ref}

\newpage

\vspace{11pt}

\vspace{11pt}

\vfill

\end{document}